\documentclass{article}
\PassOptionsToPackage{square,numbers}{natbib}

\usepackage[preprint]{neurips_2025}

\usepackage[utf8]{inputenc} 
\usepackage[T1]{fontenc}    
\usepackage{hyperref}       
\usepackage{url}            
\usepackage{booktabs}       
\usepackage{amsfonts}       
\usepackage{amsmath}        
\usepackage{nicefrac}       
\usepackage{microtype}      
\usepackage{xcolor}         
\usepackage{todonotes}
\usepackage{xcolor}

\usepackage{caption}
\usepackage{tabularx}

\usepackage{wrapfig}
\usepackage{placeins}
\usepackage{bm}

\title{DreamBoothDPO: Improving Personalized Generation using Direct Preference Optimization}

\author{%
  Shamil Ayupov\thanks{The first two authors contributed equally.} \\
  HSE University \\
  \texttt{shiayupov@edu.hse.ru} \\
  \And
  Maksim Nakhodnov$^{*}$ \\
  AIRI \\
  \texttt{nakhodnov17@gmail.com} \\
  \And
  Anastasia Yaschenko \\
  Sber AI \\
  \texttt{ASeYashchenko@sberbank.ru} \\
  \And
  Andrey Kuznetsov \\
  AIRI, Sber, Innopolis \\
  \texttt{kuznetsov@airi.net} \\
  \And
  Aibek Alanov \\
  HSE University, AIRI \\
  \texttt{alanov.aibek@gmail.com} \\
}

\begin{document}

\maketitle

\begin{abstract}
Personalized diffusion models have shown remarkable success in Text-to-Image (T2I) generation by enabling the injection of user-defined concepts into diverse contexts.
However, balancing concept fidelity with contextual alignment remains a challenging open problem.
In this work, we propose an RL-based approach that leverages the diverse outputs of T2I models to address this issue.
Our method eliminates the need for human-annotated scores by generating a synthetic paired dataset for DPO-like training using external quality metrics.
These better–worse pairs are specifically constructed to improve both concept fidelity and prompt adherence.
Moreover, our approach supports flexible adjustment of the trade-off between image fidelity and textual alignment.
Through multi-step training, our approach outperforms a naive baseline in convergence speed and output quality.
We conduct extensive qualitative and quantitative analysis, demonstrating the effectiveness of our method across various architectures and fine-tuning techniques. The source code can be found
at \href{https://github.com/ControlGenAI/DreamBoothDPO}{github.com/ControlGenAI/DreamBoothDPO}.
\end{abstract}

\section{Introduction}
Text-to-Image (T2I) diffusion models~\citep{ramesh2022hierarchical, saharia2022photorealistic, rombach2022high} have recently achieved remarkable progress, generating diverse, high-fidelity images that closely align with textual prompts. In parallel, personalization techniques\citep{DB, TI} have emerged that enable the integration of novel visual concepts into pre-trained models. However, this personalization often comes at the expense of prompt adherence, highlighting a fundamental trade-off between textual alignment and the fidelity of the injected concept. Resolving this trade-off remains a central challenge in Personalized Image Generation.

Reinforcement learning (RL) has shown promise in enhancing various aspects of T2I generation, including alignment with human preferences~\citep{ID-Aligner}, visual quality~\citep{PPOD}, diversity, and prompt fidelity~\citep{VersaT2I}. While online RL methods based on direct score optimization can significantly improve target metrics~\citep{REFL}, they frequently suffer from overfitting and diversity collapse~\citep{REFLAIG}. Conversely, Direct Preference Optimization (DPO)~\citep{rafailov2024directpreferenceoptimizationlanguage} and related methods typically require costly, manually curated datasets of paired "better–worse" examples, limiting their applicability.

In this work, we adapt the DPO framework to Personalized Generation and address its data requirements by introducing a fully automatic, customizable algorithm for pair generation. Our method exploits the intrinsic variability in outputs from T2I models and employs external scoring functions to assess concept fidelity and prompt alignment. It allows for flexible control over the trade-off between these two competing objectives.

Through systematic analysis, we optimize the computational efficiency of our approach using a multi-stage training scheme that enhances overall image quality. Extensive quantitative and qualitative evaluations, including a user study, demonstrate that our method improves Image Similarity (IS) and Text Similarity (TS)~\citep{DB} in a fully automatic setup.

Our main contributions are as follows:
\begin{itemize}
	\item We adapt DPO-style training to the personalized generation setting and propose a fully automated dataset construction pipeline.
	\item We analyze the sensitivity of key hyperparameters and identify configurations that balance performance and computational efficiency.
	\item We demonstrate the effectiveness of our approach through comprehensive experiments and user studies, showing improvements across multiple baselines and architectures.
\end{itemize}


\section{Related Work} \label{sec:related}
\textbf{Personalized Text-to-Image Generation.}
Personalized generation methods enable the injection of user-defined concepts into pre-trained diffusion models. DreamBooth~\citep{DB} and Textual Inversion~\citep{TI} pioneered this field by fine-tuning models with a unique identifier tied to reference images, enabling concept recontextualization. Subsequent works improved efficiency and multi-concept handling: SVDiff~\citep{SVDDiff} reduced parameter space through singular value decomposition, while Custom Diffusion~\citep{CD} enabled joint optimization of multiple concepts via constrained adaptation, and IP-adapter~\citep{IP-adapter}, ELITE~\citep{ELITE}, HyperDreamBooth~\citep{HDB}, BLIP-Diffusion~\citep{BLIP-Diffusion}, and Subject-Diffusion~\citep{Subject-Diffusion} earned rapid personalization using hypernetworks. While these methods reduce computational costs, they retain the fundamental fidelity-alignment trade-off inherent to concept specialization. Our work addresses this limitation by introducing an optimization framework that controllably balances fidelity and alignment without architectural modifications, leveraging automated quality metrics rather than manual regularization.

\textbf{Reinforcement Learning and Preference Optimization.}
Reinforcement learning and preference-based methods have emerged as tools for aligning diffusion models with complex objectives. ReFL~\citep{REFL}, which directly optimizes diffusion models against a reward model, showed superior performance in human evaluations on different downstream tasks, including improving human alignment, while TexForce~\citep{TexForce} and B2-DiffuRL~\citep{B2-DiffuRL} optimize prompt adherence. However, ReFL-like approaches suffer from diversity collapse and overfitting \citep{REFLAIG}. Direct Preference Optimization (DPO) \citep{DiffDPO} bypasses reward modeling but depends on manually curated preference data for complex reward functions. Several methods overcome this issue using automatic pair-dataset creation. Multiple DPO-like methods generate training data using ranking from the reward function and successfully improve in various domains in an automated setup. Particularly, ID-Aligner~\citep{ID-Aligner}, PPOD~\citep{PPOD} improve human alignment, VersaT2I~\citep{VersaT2I} improve multiple aspects ranging from text alignment to geometry. PSO~\citep{PSO} optimizes the time-distilled models for personalized generation. However, it restricts the set of "winning" images as the reference set of concept images, which limits the model's ability to adapt to diverse backgrounds and does not consider the trade-off between concept fidelity and prompt adherence. PatchDPO~\citep{patchdpo} extends DPO to personalized generation using patch-level rewards obtained from pre-trained vision models. However, PatchDPO requires expensive fine-tuning and cannot directly control the trade-off between global image metrics. In contrast, our method generates synthetic preference pairs using CLIP-based image-text alignment and concept fidelity scores, enabling fully automated training.

\section{Preliminaries}

\subsection{Diffusion Models}

In this work, we adopt conditional Latent Diffusion Models (LDMs), specifically the Stable Diffusion family~\citep{sd}, as our baseline architecture. LDMs operate in the latent space defined by a Variational Autoencoder (VAE)~\citep{vae}, where an input image $x$ is first encoded into its latent representation $z = E(x)$. A Gaussian Markovian forward diffusion process progressively adds noise to the latent representation, following the forward kernel:
$p(z_{t} \mid z_{t-1}) = \mathcal{N}(z_{t} \mid \alpha_{t}z_{t-1}, \sigma_{t}^{2}\mathbf{I}) = \alpha_{t} z_{t - 1} + \sigma_{t} \epsilon$, where $\epsilon \sim \mathcal{N}(0, \mathbf{I})$.

The reverse process is conditioned on a textual prompt $c$ and is modeled by a noise prediction network $\epsilon_{\theta}(z_{t}, c, t)$. This network is trained to minimize the variational bound on the data log-likelihood, which reduces to the following MSE objective:
\begin{equation}
    \mathcal{L}_{\text{DDPM}} = \mathbb{E}_{(x_{0}, c) \sim p(x_{0}, c), t \sim \mathcal{U}\{0, T\}, z_{t} \sim p(z_{t} | E(x_{0}))} \left[\left\| \epsilon - \epsilon_{\theta}(z_{t}, c, t)\right\|^{2}_{2}\right]
\end{equation}

Sampling begins by drawing a latent vector $z_{T} \sim \mathcal{N}(\mathbf{0}, \mathbf{I})$. This noisy latent is then iteratively denoised using DDIM sampling~\citep{ddim}:
\begin{equation}
    z_{t-1} = \text{DDIM}(z_{t-1}, \epsilon_{\theta}(z_{t}, c, t), t), t \in [T, ..., 1]
\end{equation}
Finally, the denoised latent $z_{0}$ is decoded into the output image via the VAE decoder $x_{0} = D(z_{0})$.
    
\subsection{Personalized Generation}
Personalized Generation leverages a small concept-specific dataset $\mathbb{C} = \{x_{i}\}_{i=1}^{N}$ to adapt a pretrained diffusion model for generating personalized content. To encode the concept within text prompts, a unique token $V^{\star}$ is associated with the concept through fine-tuning. The model is trained to denoise latent representations of the concept instances by minimizing the following objective:
\begin{equation}
\label{eq:loss_pg}
\mathcal{L} = \mathbb{E}_{x \in \mathbb{C}, t\sim\mathcal{U}\{0,T\},z_{t}\sim(z_{t}|E(x))}\left[\left\|\epsilon-\epsilon_\theta(z_{t}, c^{\prime}, t)\right\|_2^2\right],
\end{equation}
where the conditioning prompt $c^{\prime}=$\textit{"a photo of a V*"} includes the special token to guide generation toward the target concept.

\subsection{Direct preference optimization (DPO)}
DPO~\citep{DiffDPO} is a method for aligning generative models with human preferences without relying on reinforcement learning. Unlike RLHF~\citep{rlhf}, which trains a separate reward model and optimizes outputs via policy gradients, DPO directly adjusts the model using pairwise preference data.
Given a context $c$ and pair of "better-worse" images $x_{w} \succ x_{l}$, DPO minimizes the following loss:
\begin{equation}
\label{eq:loss_dpo}
    \footnotesize L_{\text{DPO}}(\theta)\!=\!-
    \mathbb{E}_{c,x_w,x_l}\left[
    \log\sigma\left(\beta \log \frac{p_{\theta}(x_w|c)}{p_{\text{ref}}(x_w|c)}-\beta \log \frac{p_{\theta}(x_l|c)}{p_{\text{ref}}(x_l|c)}\right)\right],
\end{equation}
where $p_{\theta}$ is the fine-tuned model, $p_{\text{ref}}$ is the reference model, and $\beta$ controls regularization. This approach directly steers the model toward preferred outputs using maximum likelihood principles, bypassing the complexity of reward modeling.

\section{Method}
\label{sec:method}

\subsection{Motivation}
\label{sec:motivation}

\begin{figure}[tb]
    \centering
    \includegraphics[width=\linewidth]{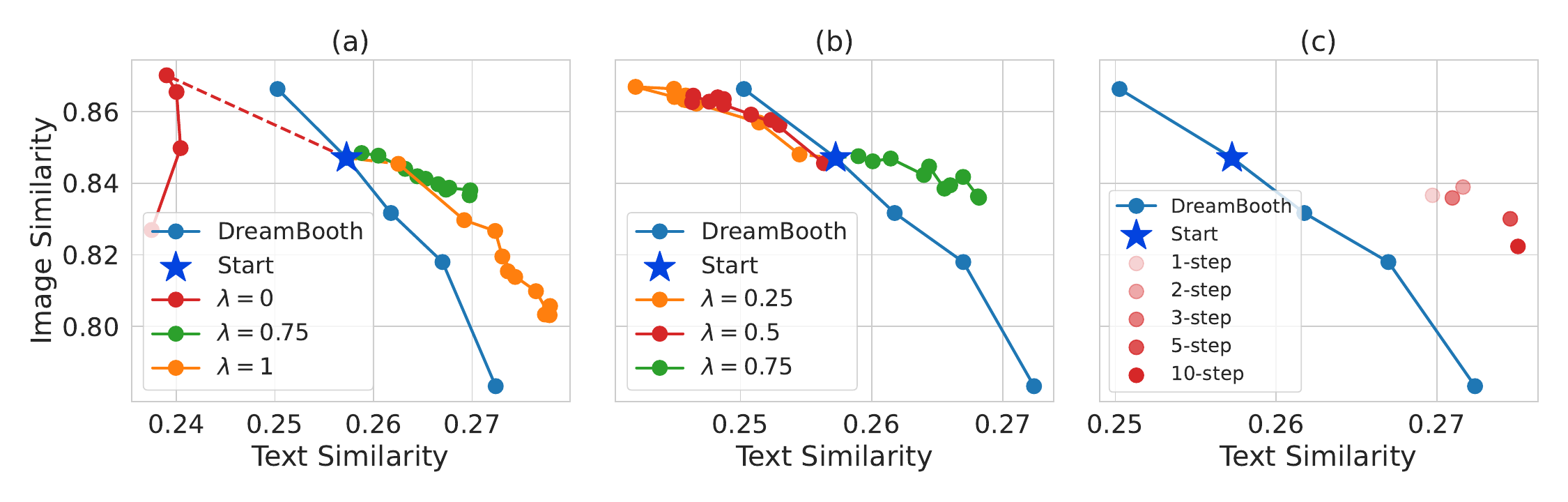}
    \caption{(a) Individual optimization of the IS ($\lambda = 0$) and TS ($\lambda = 1$) allows to improve the target metric but drastically degrades the other one. (b) The weighted combination allows stabilized training but remains oversensitive to the weighting coefficient. (c) Multistep training can lead to significant improvements; however, it lacks effective directional control.}
    \label{fig:ablation-1}
\end{figure}

Although many fine-tuned personalization algorithms like DreamBooth can provide impressive concept fidelity, their textual quality often degrades significantly during optimization. Overall, the method forms a frontier that defines an optimal trade-off between prompt alignment and concept fidelity as shown in Figure~\ref{fig:ablation-1}(a). Different training-time and inference-time techniques can shift this frontier\citep{realc},~\citep{persongen},~\citep{photoswap},~\citep{regfree}. Nevertheless, it is challenging to push out this Pareto curve. 

We are encouraged to use DPO since it has shown great success in multiple T2I generation tasks, including advancing both conflicting metrics~\citep{VersaT2I}, like aesthetic and prompt adherence. The main obstacle to directly implementing this approach in a personalized generation is a lack of a human-annotated paired dataset. Since the dataset must reflect the characteristics of the specific object, it cannot be collected beforehand. We can utilize the diverse generation capabilities of the baseline personalization model, as it can produce samples of varying quality in both aspects of personalization.

The required pairs can be selected manually from a large enough generation of the fine-tuned model. However, such a labor-incentive approach could not be scaled. Therefore, we consider an automatic pair selection process that will allow us to use DPO algorithms to enhance the baseline model directly. 

\subsection{Naive scoring}
\label{sec:naive}

Assume that each sample can be evaluated using a scalar scoring function $\mathcal{S}$. This function serves to differentiate high-quality image samples $x_{w}$ from lower-quality ones $x_{l}$ by comparing their scores: $\mathcal{S}(x_{w}) > \mathcal{S}(x_{l})$. In practice, however, some samples -- such as visually similar or near-duplicate images -- may receive nearly identical scores. Despite this, DPO treats all pairs equally, regardless of the magnitude of the score difference, applying the same learning signal to both clearly distinguishable and marginally different pairs.
To address this, we introduce a filtering mechanism that distinguishes significant from insignificant preference pairs. Specifically, we retain only those pairs where the score gap exceeds a predefined threshold $\tau$: $\mathcal{S}(x_{w}) - \mathcal{S}(x_{l}) > \tau$. This filtering ensures that DPO focuses on more informative comparisons, leading to a more stable and meaningful optimization process.

\begin{figure}[b]
    \centering
    \includegraphics[width=\linewidth]{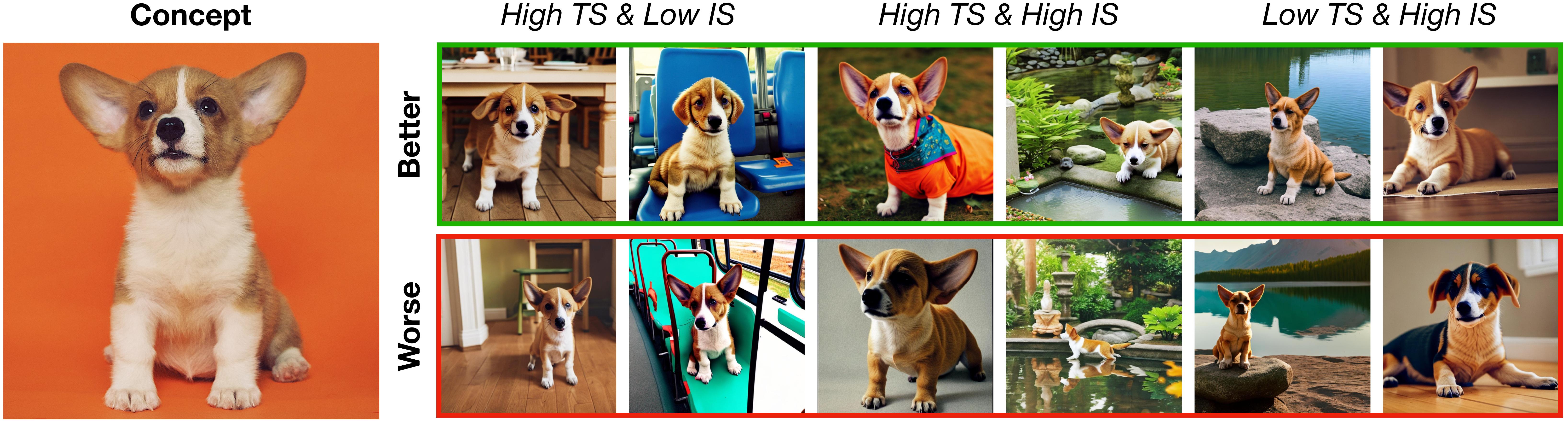}
    \caption{Pairs of images with different IS/TS balances. The prompts for the columns from left to right are: \textit{"a $V^{\star}$ sitting beneath table and chairs"}, \textit{"a $V^{\star}$ is sitting underneath a seat on a bus"}, \textit{"a $V^{\star}$ wearing a traditional sari"}, \textit{"a $V^{\star}$ in a serene Zen garden with koi ponds"}, \textit{"a $V^{\star}$ sitting on a rock area at water's edge of a lake"}, \textit{"a $V^{\star}$ laying on the floor chewing on a stick".}
    }
    \label{fig:pairs}
\end{figure}

\begin{figure*}[t]
  \centering
      \includegraphics[clip,width=\linewidth]{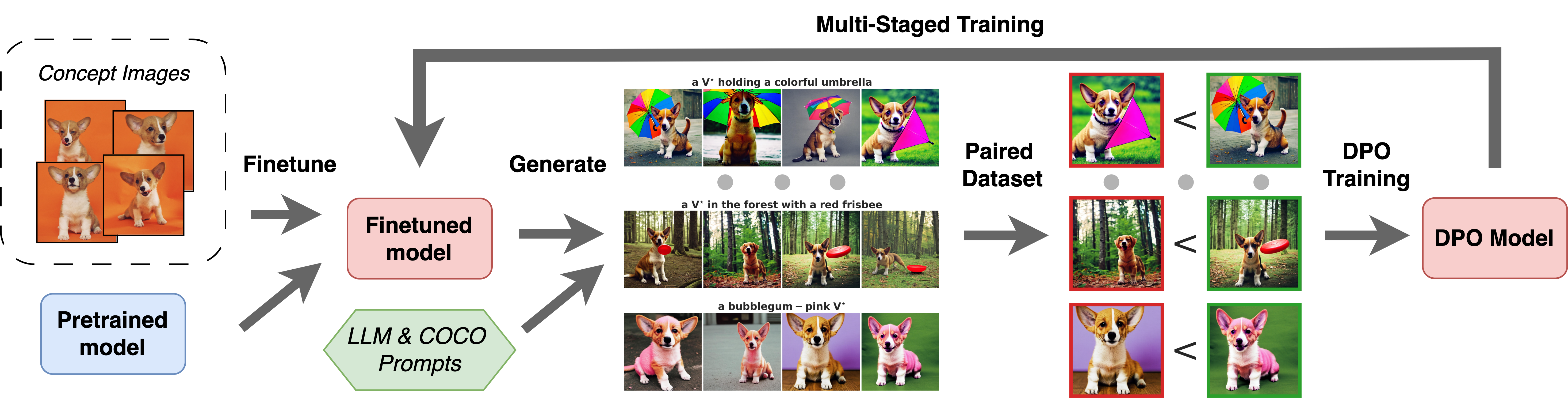}
      \captionof{figure}{An outline of the proposed method. First, we fine-tune the personalized model and generate a diverse set of images, capturing the model's output variability. Then, images are scored and used to create a paired dataset for DPO training. The process can be repeated to form a multi-step training.}
      \label{fig:method}
\end{figure*}

The overall training pipeline is illustrated in Figure~\ref{fig:method}. Starting with a reference set of concept images $\mathbb{C}$, we first fine-tune a pre-trained diffusion model for personalized generation using Equation~\ref{eq:loss_pg}. Next, we sample a diverse set of $N$ textual prompts and generate $M$ images per prompt to capture a wide range of generation outcomes, including both high- and low-quality samples. This variability is essential for constructing informative "better–worse" preference pairs.

We then apply the pair selection strategy detailed in the following sections to filter and select meaningful preference pairs for training. These pairs form the dataset used in the DPO fine-tuning stage, which adjusts the model to better align with the desired trade-offs between concept fidelity and prompt adherence. This process can be repeated iteratively -- regenerating samples, re-evaluating scores, and updating training pairs -- forming the multi-step training regime described in Section~\ref{sec:multi}.

The second design choice concerns the source of images to be scored. To ensure that we can generalize beyond the training set, we utilize a variety of prompts: $3000$ captions from the COCO dataset~\citep{coco} and $1000$ generated by a large language model~\citep{chatgpt}. This variety encourages broader image generation and increases the likelihood of capturing meaningful quality differences. The data collection process is outlined in Section~\ref{sec:setup} and in Appendix~\ref{app:data}.

The most straightforward way to define a scoring function is by leveraging the well-established CLIP-based~\citep{clip} Image and Text Similarities:
\begin{equation}
 \text{IS}(x) = \frac{1}{|\mathbb{C}|} \sum_{x_i \in \mathbb{C}} \cos(\text{CLIP-I}(x), \text{CLIP-I}(x_i)), \quad \text{TS}(x) = \cos(\text{CLIP-I}(x), \text{CLIP-T}(c)),
 \end{equation}
where, $\text{IS}(x)$ measures the similarity of the generated image $x$ to a set of reference images $\mathbb{C}$, while $\text{TS}(x)$ evaluates alignment with the target prompt $c$. Although optimizing for $\text{IS}$ or $\text{TS}$ individually does improve the corresponding metric, Figure~\ref{fig:ablation-1}(a) shows that the overall trade-off remains comparable to -- or worse than -- the DreamBooth frontier.

To mitigate this issue, one can combine the two metrics into a single score:
\begin{equation}
 \mathcal{S}(x) = \lambda \cdot \text{TS}(x) + (1 - \lambda) \cdot \text{IS}(x)
 \end{equation}
As shown in Figure~\ref{fig:ablation-1}(b), this weighted combination helps reduce image fidelity degradation while significantly improving prompt adherence. However, this method offers limited control over the trade-off between the two objectives and is sensitive to the choice of the weighting parameter $\lambda$.

\subsection{Multistep Training}\label{sec:multi}
\label{sec:multistep}

A single-step DPO training assumes that the model remains static throughout the process. In practice, the model evolves during training, and the quality distribution of generated samples changes. Therefore, pairs selected using the initial model may no longer reflect meaningful preferences as training progresses. This mismatch can degrade performance, particularly when the score function fails to capture the updated trade-off between prompt adherence and image fidelity.

To address this issue, we adopt a multistep training procedure. At each step, we generate a fresh set of samples using the current version of the model, re-score them, and re-select preference pairs based on the updated score distribution. This allows the training process to adapt dynamically to the model’s evolving behavior.

We experiment with varying the number of DPO training steps: $1$ (single-step), $2$, $3$, $5$, and $10$. Figure~\ref{fig:ablation-1}(c) shows that using $2$-$3$ steps already improves performance over the single-step baseline. Increasing the number of steps to $5$ or $10$ further boosts prompt adherence but at the cost of noticeable degradation in image fidelity. This suggests that while multistep training helps to better align with the text prompt, it may also push the model too far from the reference distribution.

Overall, multistep training proves beneficial, but it introduces a new challenge: directional control. Because we rely solely on scalar scores to select pairs, the training signal can be biased by poorly aligned pairs -- that is, preference pairs where the "improvement" lies in a direction that harms other aspects of generation quality. We explore this issue in detail in the next section through an angle-based analysis of score dynamics.

\subsection{Score Analysis}
\label{sec:scores}

\begin{figure*}[tb]
  \centering
      \includegraphics[clip,width=\linewidth]{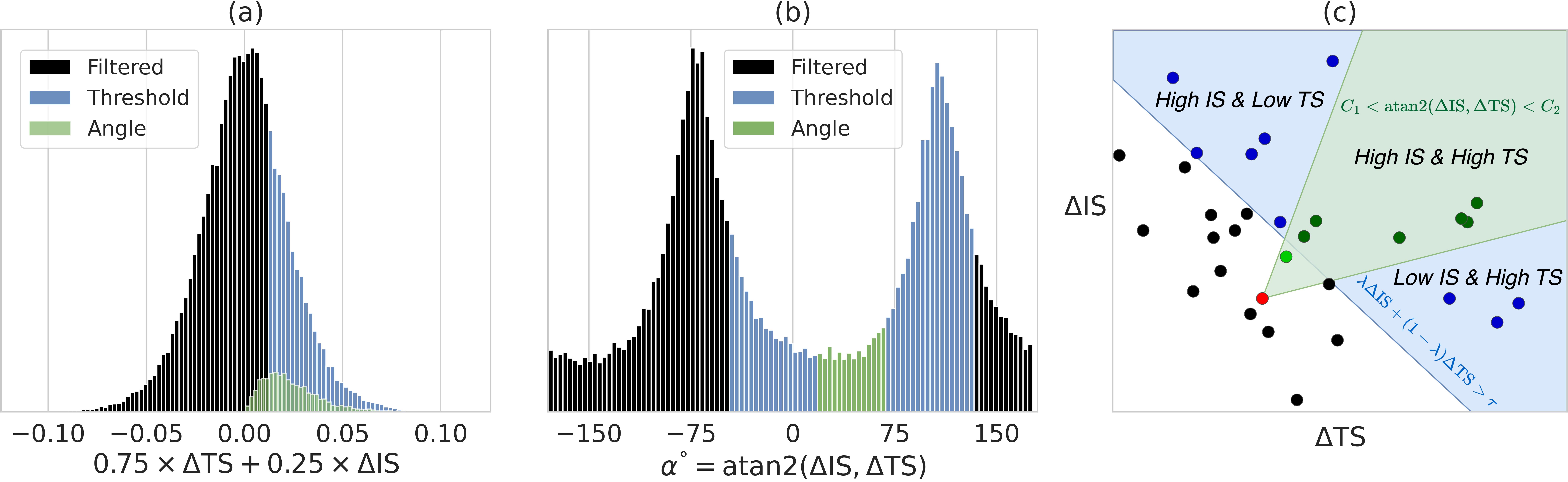}
      \captionof{figure}{(a) The distribution of the weighted score function for different samples exhibits an unimodal behavior, failing to capture the \textit{High TS \& High IS} region. (b) The distribution of angles shows that filtering by threshold fails to remove pairs from both modes, while angle filtering can separate the required region. (c) Depiction of possible pairs for one selected (red) sample. While threshold filtering captures harmful samples from \textit{High TS \& Low IS} and \textit{Low TS \& High IS} regions, angle filtering selects a small fraction of pairs with high positive signals.}
      \label{fig:angles_hist}
\end{figure*}

\begin{figure}[tb]
    \centering
    \includegraphics[width=\linewidth]{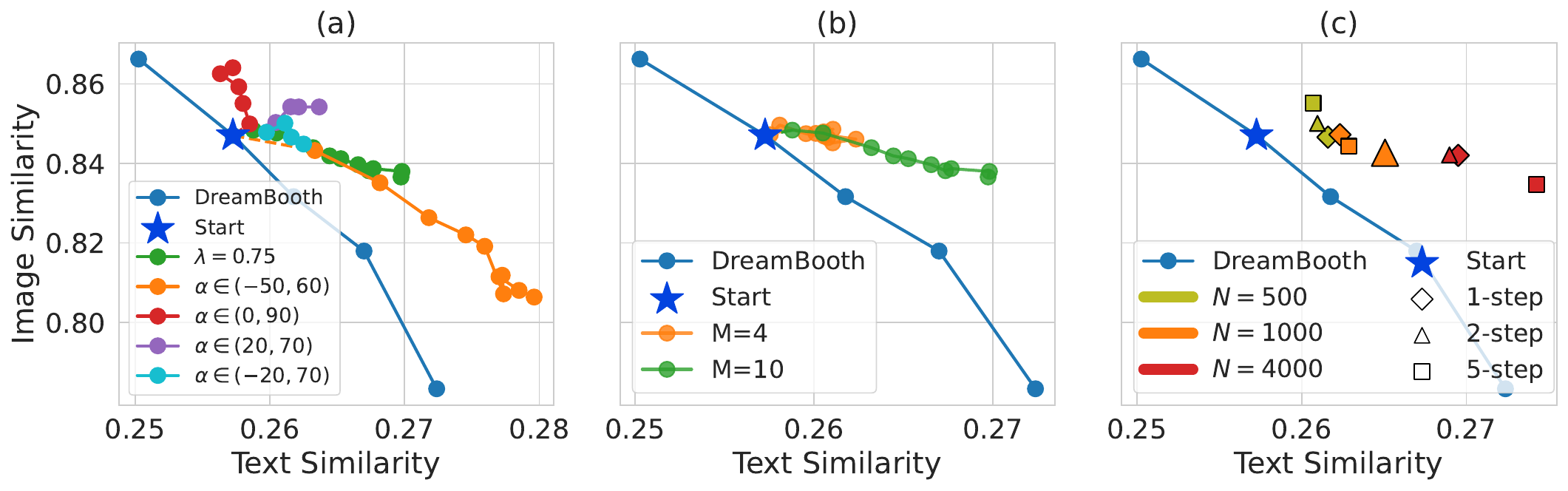}
    \caption{(a) Angle-based filtering allows for finer directional control. (b) Reducing the number of images per prompt negatively affects the performance. (c) 2-step $N=1000$ (large triangle) setup improves performance of 1-step setup and lowers computational costs of $N=4000$.}
    \label{fig:ablation-2}
\end{figure}

Thresholding preference pairs by the scalar score $\mathcal{S}(x)$ is a coarse selection strategy -- it considers only the magnitude of improvement, not its direction. As training progresses, this can lead to degenerate solutions, such as optimizing for prompt adherence while severely degrading image quality.

To better understand and guide optimization, we analyze training pairs in the $(\Delta \text{TS}, \Delta \text{IS}) = (\text{TS}(x_{w}) - \text{TS}(x_{l}), \text{IS}(x_{w}) - \text{IS}(x_{l}))$ space. Each pair is associated with an angle $\alpha = \text{atan2}(\Delta \text{IS}, \Delta \text{TS})$, which indicates the direction of improvement across the two metrics.
Figure~\ref{fig:angles_hist}(3) illustrates that threshold-based filtering (blue) selects pairs from a wide range of directions, including extremes that may harm one metric. Examples of the pairs from each region are presented in Figure~\ref{fig:pairs}.

To address this, we propose angle-based filtering (green), which selects pairs satisfying:
 \begin{equation}
 C_{1} < \text{atan2}(\Delta\text{IS}, \Delta\text{TS}) < C_{2}
 \end{equation}
This constraint ensures updates lie within a controlled region of the trade-off space, favoring improvements in both TS and IS or at least avoiding significant degradation in either.

Threshold-based filtering is a special case of this approach: when $\tau = 0$, it corresponds to an angular cone of $180^\circ$. However, as shown in Figure~\ref{fig:angles_hist}(c), thresholding can include conflicting directions (e.g., \textit{High IS \& Low TS} vs. \textit{Low IS 
\& High TS}), which may cancel each other out. This limits its ability to target the desirable region of joint improvement (\textit{High TS \& High IS}). In contrast, angle-based filtering leverages the bimodal structure of the angle distribution (Figure~\ref{fig:angles_hist}(b)), unlike the unimodal score distribution in Figure~\ref{fig:angles_hist}(a), enabling more selective and effective pair sampling.

This directional filtering allows finer control over the optimization trajectory. Adjusting the bounds $C_1$ and $C_2$ modulates the emphasis on each metric: increasing $C_2$ favors IS, while decreasing $C_1$ favors TS. We define three base setups: -TS with $(C_1, C_2) = (-20, 70)$ directed to improve TS, -IS with $(C_1, C_2) = (0, 90)$ for IS, and -MIX with $(C_1, C_2) = (-10, 80)$ that balances both metrics. As shown in Figure~\ref{fig:ablation-2}(a), DreabBoothDPO enables precise steering of the trade-off, preventing collapse into one extreme and promoting balanced model behavior. 

\subsection{Training Analysis}
\label{sec:analysis}

While the method offers strong controllability and can extend the Pareto frontier, it comes with a high computational cost. The image generation phase scales linearly with the number of prompts and images produced per prompt, as multiple samples are needed to leverage the model’s intrinsic variability. To lower this cost, we compare configurations with $10$ and $4$ images per prompt. As shown in Figure~\ref{fig:ablation-2}(b), although the optimization trajectory remains similar, the overall performance gain drops substantially when using fewer images. We attribute this to the fact that the number of candidate preference pairs grows quadratically with the number of images per prompt, while generation cost increases only linearly.

To improve efficiency, we shift focus to the second key hyperparameter: the number of prompts. Figure~\ref{fig:ablation-2}(c) demonstrates that multi-step training can close the performance gap between a more expensive setup with 4000 prompts and a more efficient one with 1000 prompts. Based on this finding, we adopt a configuration with 1000 prompts and 10 images per prompt, trained using two-step DPO, as our base model for all subsequent experiments.

\section{Experiments}
\label{sec:experiments}

\begin{figure*}[t]
  \centering
  \begin{minipage}{.55\textwidth}
      \centering
      \includegraphics[clip,width=\linewidth]{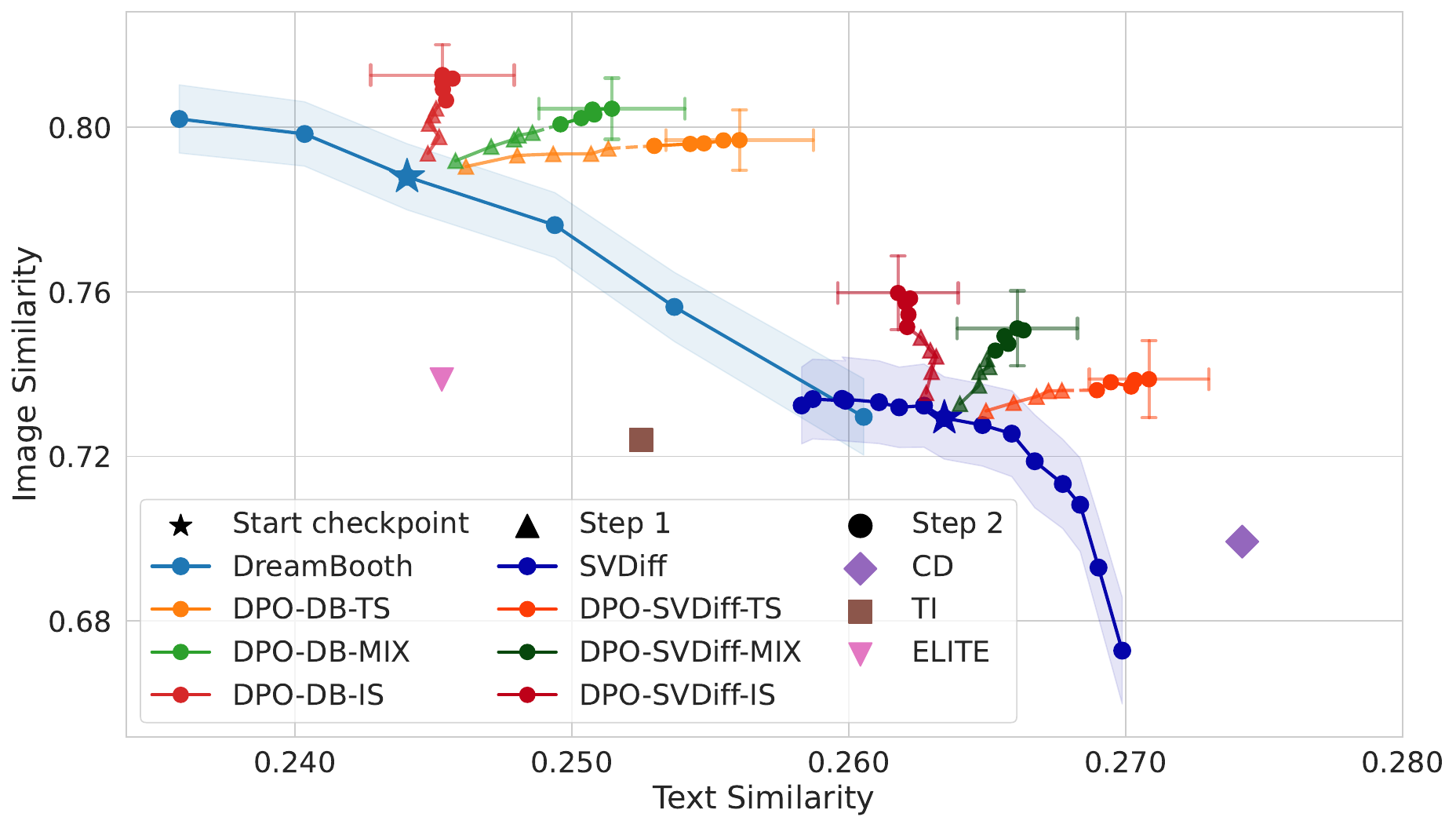}
      \captionof{figure}{SD2 results across DPO variants. -TS setup improves prompt alignment, -IS enhances visual similarity, and -MIX balances both. All variants outperform baselines, enabling controllable trade-offs of TS and IS.}
      \label{fig:main:sd2}
  \end{minipage}
  \hfill
  \begin{minipage}{.40\textwidth}
      \centering
      \includegraphics[clip,width=\linewidth]{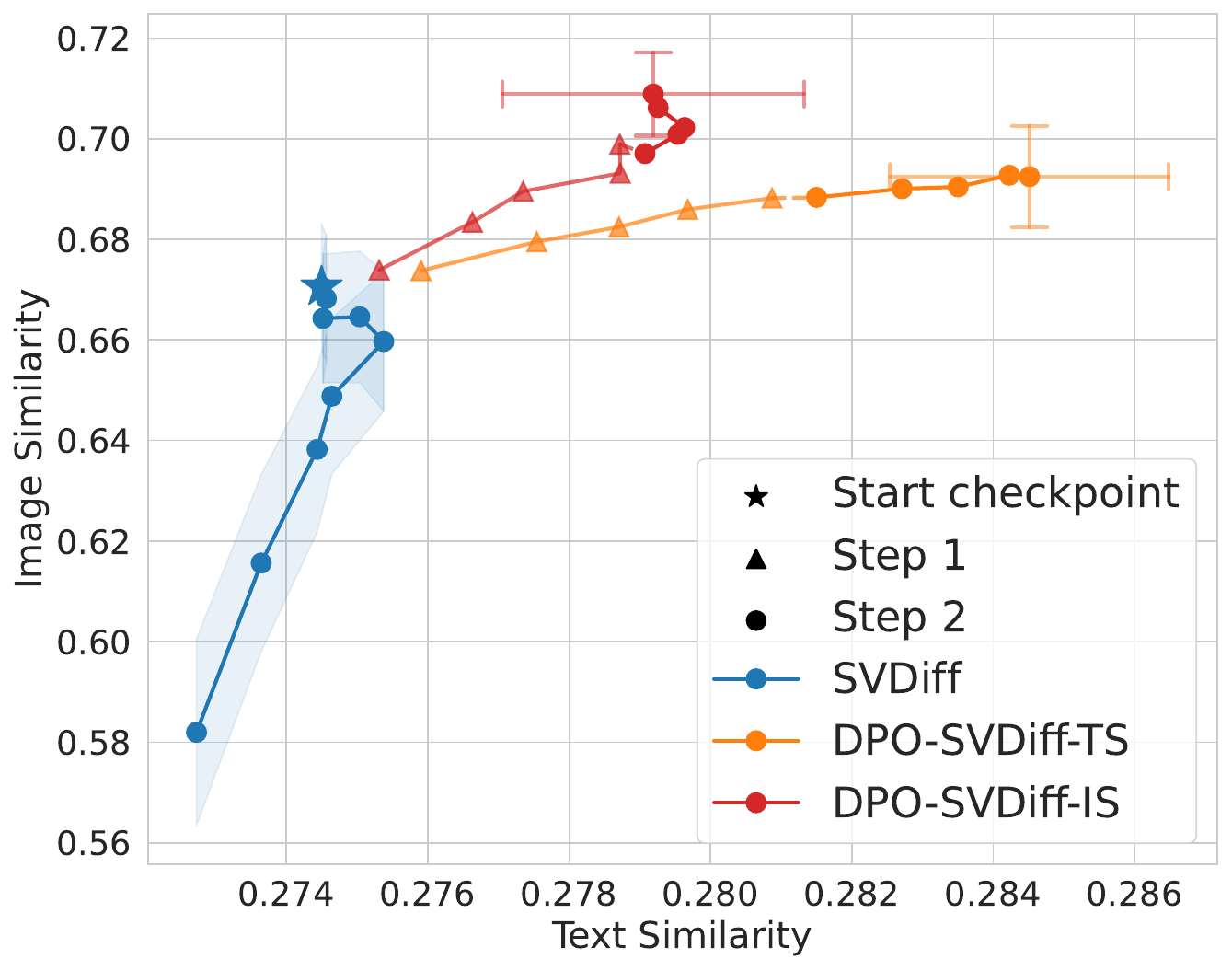}
      \captionof{figure}{Performance of DB-DPO on SDXL. Improvements are consistent with SD2, demonstrating strong generalization to more capable base models.}
      \label{fig:main:sdxl}
  \end{minipage}
  \label{fig:main}
\end{figure*}

\subsection{Setup}\label{sec:setup}

\paragraph{Dataset}
For each training concept, we collect a total of $3000$ concept-specific prompts using concept-category mappings from the COCO dataset~\citep{coco} and $1000$ concept-agnostic prompts generated with ChatGPT~\citep{chatgpt}. For each prompt, we generate $M=10$ samples to ensure sufficient diversity. Full details of the data collection process can be found in Appendix~\ref{app:data}.

\paragraph{Base models}

We use SD2 and SDXL as the base models. For SD2, we use DreamBooth and SVDiff as the base personalized generation methods. For SDXL we only use SVDiff as DreamBooth requires full fine-tuning, which is not feasible to train on a single GPU. We train base personalized generation methods on $30$ concepts from the DreamBench. To generate samples, we use the  PNDM scheduler~\citep{PNDM} with $50$ steps and a guidance scale of $7.5$ for SD2 and the EulerDiscrete scheduler~\citep{EDM} with $50$ steps and a guidance scale of $5.0$ for SDXL.

\paragraph{Training}

For SD2, we fine-tune all layers of the U-Net~\citep{unet} initialized with either DreamBooth or SVDiff checkpoints. We use a DPO regularization coefficient $\beta=5000$, learning rate of $2.5\times10^{-6}$, and batch size of $256$. For SDXL~\citep{sdxl}, we apply LoRA-based~\citep{lora} fine-tuning with rank $4$. We use $\beta=5000$, learning rate of $6.4\times10^{-5}$, and batch size of $64$. The number of training steps is chosen such that each preference pair is seen $5$ times -- this setup yields a good balance between convergence and generation quality. In Section~\ref{sec:multi}, we vary the number of steps $S \in {1, 2, 3, 5, 10}$ and find that $S=2$ provides faster convergence than $S=1$ and better quality than $S > 2$. We report other training details and analysis of the computational trade-offs of different hyperparameter selection in Appendix~\ref{app:training}.

\begin{figure}[tb]
    \centering
    \includegraphics[width=\linewidth]{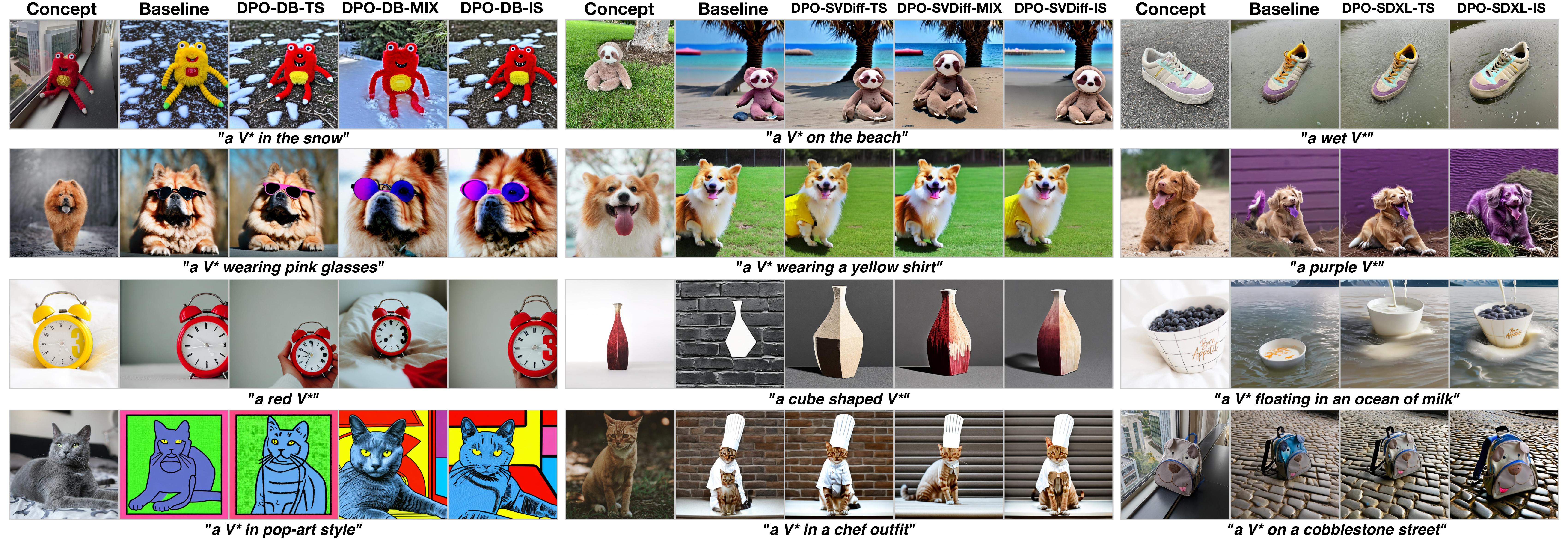}
    \caption{Qualitative examples from models fine-tuned toward different objectives. -TS setup consistently improves prompt adherence, while -IS improves concept fidelity across all setups.}
    \label{fig:samples:sd2-db}
\end{figure}

\begin{figure}[t]
    \centering
    \includegraphics[width=\linewidth]{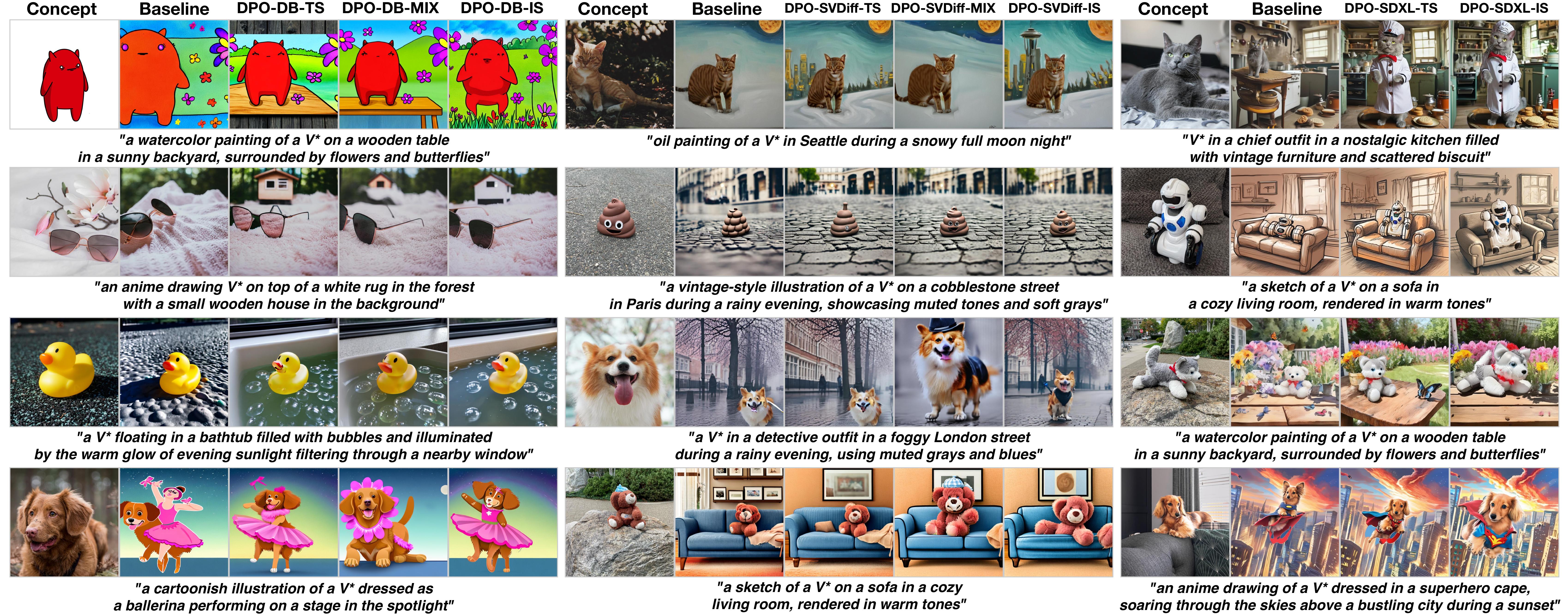}
    \caption{Qualitative examples for long prompt setup. -TS setup consistently improves prompt adherence, while -IS improves concept fidelity across all setups.}
    \label{fig:samples:long}
\end{figure}

\paragraph{Evaluation}
We evaluate the models using CLIP-I and CLIP-T scores on DreamBench prompts. We use the `live` subset for live concepts and the `object` subset for all others. Scores are averaged over 10 generated images per prompt and then across the corresponding subset of prompts. Additional qualitative and quantitative results on long and complex prompts are included in Appendix~\ref{app:long}.

\subsection{Results}

Figure~\ref{fig:main:sd2} shows that our method not only improves both Image Similarity (IS) and Text Similarity (TS) simultaneously -- surpassing Pareto frontier -- but also allows for controllable trade-offs using angle-based pair filtering. Specifically, DPO-TS significantly boosts TS while also slightly improving IS. Conversely, DPO-IS improves IS without degrading prompt relevance. The mixed variant (DPO-MIX) offers a balanced trade-off. This control is consistent across personalization backbones: DreamBooth-based models follow the same relative ranking of the -TS, -IS, and -MIX variants. As shown in Figure~\ref{fig:main:sdxl}, the method also generalizes well to more capable models such as SDXL.

Figures~\ref{fig:samples:sd2-db},~\ref{fig:samples:long} provide qualitative samples demonstrating improvements in personalization across different optimization directions. The -TS variant improves prompt adherence while maintaining concept fidelity, whereas -IS enhances the resemblance between the reference and generated object. Increasing the proportion of high-IS training pairs steadily improves visual similarity, matching the upper threshold $C_2$ used in angle-based filtering. Additional visual results are available in Appendix~\ref{app:more}.

\subsection{User Study} 
To further validate our observations, we conducted a user study. We collected $6000$ responses from $10$ expert assessors on side-by-side comparisons between DB-DPO outputs and corresponding baselines. As summarized in Table~\ref{tab:user_study}, users preferred DB-DPO over baseline outputs in a majority of cases, with up to 50\% higher preference, aligning with both IS and TS objectives. Full details of the user study protocol are included in Appendix~\ref{app:user}.

\begin{wraptable}{r}{0.44\textwidth}
\vspace{-1.5em}
\centering
    \caption{We report a percentage of total user votes where DB-DPO was preferred (Win), performed worse (Lose), or showed no noticeable difference (No Diff) compared to the baseline.}
    \label{tab:user_study}
    \resizebox{\linewidth}{!}{
        \begin{tabular}{lccccccc}
            \toprule
            & \multicolumn{3}{c}{\textbf{DPO-SVDiff-TS}} & \multicolumn{3}{c}{\textbf{DPO-SDXL-TS}} \\
            \cmidrule(lr){2-4} \cmidrule(lr){5-7}
                & Win  & Lose & No Diff   & Win  & Lose & No Diff  \\
            \midrule
            TS  &  $\bm{6.6}$ &  $3.9$ & $89.5$       & $\bm{13.5}$ & $6.8$  & $79.7$     \\
            IS  & $\bm{31.7}$ & $28.2$ & $40.1$       & $\bm{33.1}$ & $23.9$ & $43.0$     \\
            All & $\bm{23.9}$ & $20.1$ & $56.0$       & $\bm{31.7}$ & $20.5$ & $47.8$     \\
            \bottomrule
    \end{tabular}
    }
\end{wraptable}

\section{Limitations}
\label{sec:limitations}
While our multi-step regimen is already more efficient than naïve single-step DreamBoothDPO fine-tuning, the need to generate diverse candidate sets and iterate through several optimization rounds still introduces a noticeable overhead; exploring smarter data-reuse, partial sampling, or lightweight distillation could make training even leaner. Moreover, the current score functions rely on CLIP-style similarity metrics; incorporating richer, task-aware evaluations from more advanced vision–language models may provide finer guidance and further boost alignment and fidelity.

\section{Conclusion}
We introduced DreamBooth DPO, an end-to-end personalization framework that automatically assembles “better–worse” training pairs, rigorously analyzes how each stage -- from pair generation and filtering to update scheduling -- affects model quality, and, drawing on these insights, streamlines the process into a lean multi-step fine-tuning scheme. A novel directional-control mechanism lets practitioners steer optimization toward stronger concept fidelity or tighter prompt alignment without restarting training. Evaluated on diverse DreamBench concepts and multiple Stable Diffusion backbones, this controlled optimization consistently exceeds the conventional Pareto frontier of image- and text-similarity metrics, positioning DreamBooth DPO as a practical, scalable, and user-controllable foundation for personalized diffusion.


\newpage
\bibliography{references}
\bibliographystyle{ieeetr}


\newpage
\appendix


\section{Data collection}
 \label{app:data}
To ensure sufficient diversity in visual contexts and styles, we collect a total of $4000$ prompts per concept: $3000$ concept-specific prompts derived from the COCO dataset~\citep{coco} and $1000$ concept-agnostic prompts generated using a large language model~\citep{chatgpt}.

\paragraph{Concept-Specific Prompts.}
 To generate meaningful COCO-based prompts, we first filter the dataset for samples whose labeled category matches the target concept class. For instance, we select entries from the \textit{animal/dog} category for the \textit{dogX} concept corresponding to the \textit{dog} class. Some concept-to-category mappings are non-trivial; a full mapping between DreamBench concept classes and COCO categories is provided in Table~\ref{tab:concept-coco-map}.
Next, we parse the captions associated with the selected samples and retain only those containing a single, unambiguous occurrence of the target category (e.g., \textit{"a black \textbf{dog}"} is accepted, whereas \textit{"a black \textbf{dog} and a white \textbf{dog}"} or \textit{"black \textbf{dogs}"} are not). We then replace the category word with the placeholder token [V*] to indicate the concept position. From this filtered set, we sample $3000$ unique prompts per concept.

\paragraph{Concept-Agnostic Prompts.}
Although COCO prompts describe realistic scenes, they can be of low quality. To address this issue, we supplement the dataset with concept-agnostic prompts generated using an LLM. We define five prompt categories inspired by DreamBench: \textit{appearance}, \textit{outfit}, \textit{background}, \textit{style}, and \textit{position}. For each category, we curate a few seed examples (e.g., \textit{"a [V*] in anime style"} for \textit{style}, or \textit{"a [V*] in a cave"} for \textit{background}) and prompt the LLM to generate $200$ similar but distinct examples per category. After generation, we manually review the prompts to eliminate duplicates and low-quality results, yielding a final set of $1000$ diverse, concept-agnostic prompts. These prompts are shared across all concepts. Examples of collected prompts for a specific concept are shown in Table~\ref{tab:prompts-examples}.

\begin{table}[b]
\begin{minipage}{.42\textwidth}
    \centering
    \begin{tabular}{ll}
        Concept class & COCO category/class \\\hline
        backpack & accessory/backpack \\
        boot & accessory/* \\
        bowl & kitchen/bowl \\
        can & kitchen/bottle \\
        candle & indoor/* \\
        cartoon & indoor/teddy\_bear \\
        cat & animal/cat \\
        clock & indoor/clock \\
        dog & animal/dog \\
        glasses & accessory/* \\
        sneaker & accessory/* \\
        stuffed animal & indoor/teddy\_bear \\
        teapot & kitchen/bottle \\
        toy & indoor/teddy\_bear \\
        vase & indoor/vase \\
    \end{tabular}
    \vspace{0.6em}
    \caption{Correspondence between concept classes and COCO prompts}
    \label{tab:concept-coco-map}
\end{minipage}
\hfill
\begin{minipage}{.54\textwidth}
    \centering
    \begin{tabular}{ll}
        Source & Prompt \\\hline
        COCO & a [V*] sits on top of a bed  \\
        COCO & a small [V*] laying on \\
        & \;\;a bean bag \\
        COCO & a collared [V*] playing with  \\
        & \;\;a toy on a grassy lawn \\
        COCO & large white wet [V*] sitting \\
        & \;\; in the shower \\
        COCO & a [V*] that is holding \\
         & \;\;a bowl in his mouth \\
        LLM (appearance) & a neon-lit [V*] \\
        LLM (outfit) & a [V*] in a biker jacket \\
        LLM (background) & a [V*] with windmills \\
         &  \;\;on the horizon \\
        LLM (style) & a [V*] in steampunk style \\
        LLM (position) & a [V*] riding a scooter\\
    \end{tabular}
    \vspace{0.6em}
    \caption{Examples of the collected prompts for the \textit{dog} concept class}
    \label{tab:prompts-examples}
\end{minipage}
\end{table}
\section{Training details} \label{app:training}
\begin{figure*}[t]
  \centering
  \begin{minipage}{.49\textwidth}
      \centering
      \includegraphics[clip,width=\linewidth]{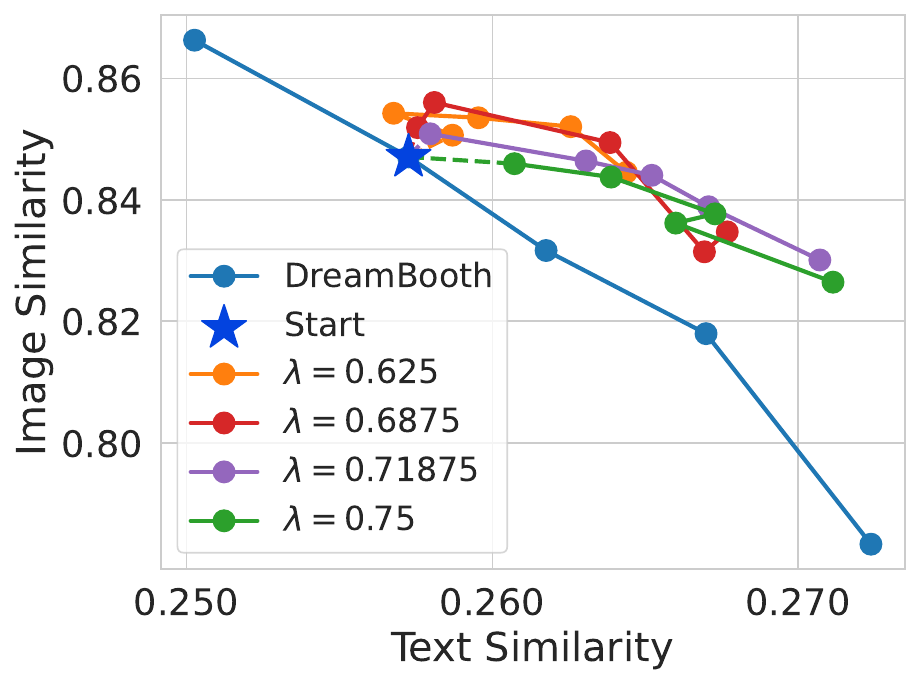}
      \captionof{figure}{DB-DPO trajectories across several fine-grained $\lambda$ on a $500$-prompt subset for \textit{dog6} concept. No significant changes are observed.}
      \label{fig:appendix:fine-lambda}
  \end{minipage}
  \hfill
  \begin{minipage}{.49\textwidth}
      \centering
      \includegraphics[clip,width=\linewidth]{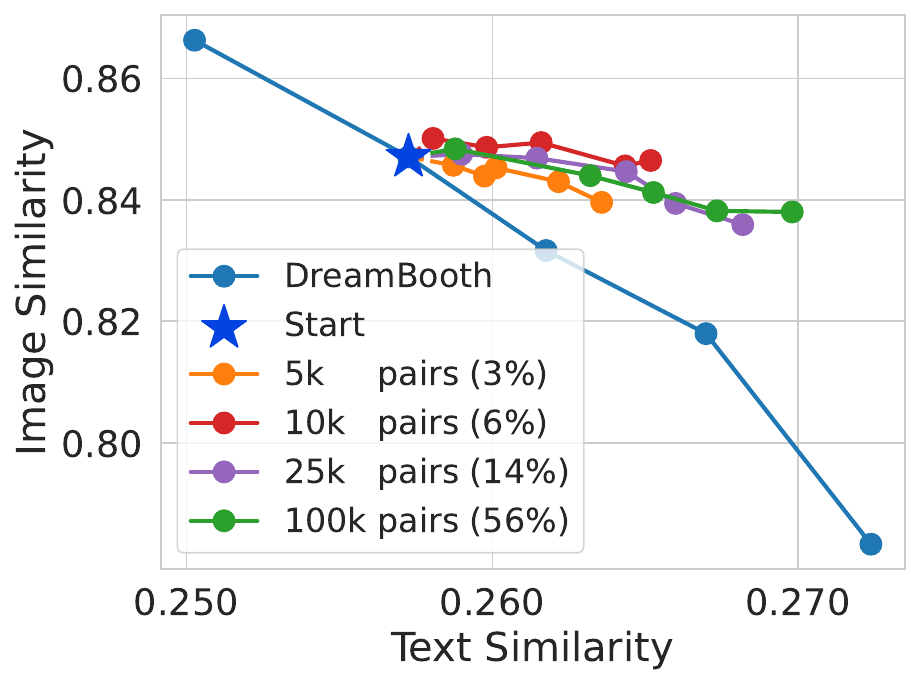}
      \captionof{figure}{DB-DPO trajectories with varying $\tau$ thresholds on a $4000$-prompt subset for the \textit{dog6} concept. Only minor changes are observed.}
      \label{fig:appendix:thresholds}
  \end{minipage}
\end{figure*}

\paragraph{Additional sweeps}  
In addition to the coarse sweep over $\lambda \in \{0, 0.25, 0.5, 0.75, 1\}$, we also explore a finer set of values: $\lambda \in \{0.625, 0.6875, 0.71875\}$. However, this finer sweep does not yield notable changes in the trajectory's behavior, as shown in Figure~\ref{fig:appendix:fine-lambda}. The lack of clear trends, combined with the oversensitivity observed in the coarse sweep, suggests that score-based filtering lacks robustness.
We also experimented with varying the threshold $\tau$ in score-based filtering, retaining $\{56\%, 14\%, 6\%, 3\%\}$ of the top-scoring pairs. Figure~\ref{fig:appendix:thresholds} shows that this did not result in significant changes in the training trajectory and instead introduced an unnecessary slowdown. As a result, we shifted our focus to angle-based filtering and did not further pursue score-based approaches.

\paragraph{Runtime analysis}
The total time of a pipeline consists of time to generate samples and to fine-tune the model. On \textit{Nvidia H100} GPU generation requires $\sim$ $1$ hour to generate $10000$ images. This motivates us to reduce the number of samples. The full setup with $4000$ prompts and $M=10$ requires $4$ hours, while the main setup with $1000$ prompts requires only $1$ hour. The main setup is trained for $500$ steps, which takes $\sim$ $1.5$ hours. Thus, the full training pipeline takes $2.5$ hours total.

\paragraph{Evaluation details}
The evaluation protocol is as follows. For a given concept and a prompt, we generate $10$ images. We average the metrics (CLIP-I, CLIP-T) over the generated images, and over concept images for CLIP-I. We repeat this for all prompts from the corresponding set of prompts (\textit{live} for live concepts -- \textit{cat} and \textit{dog}, and \textit{object} for others) and average the results. Finally, we do this over all the concepts and report the average as well as $1\sigma$ error intervals.

\section{Long prompts evaluation} \label{app:long}
\begin{figure*}[b]
\vspace{-1em}
  \centering
  \begin{minipage}{.55\textwidth}
      \centering
      \includegraphics[clip,width=\linewidth]{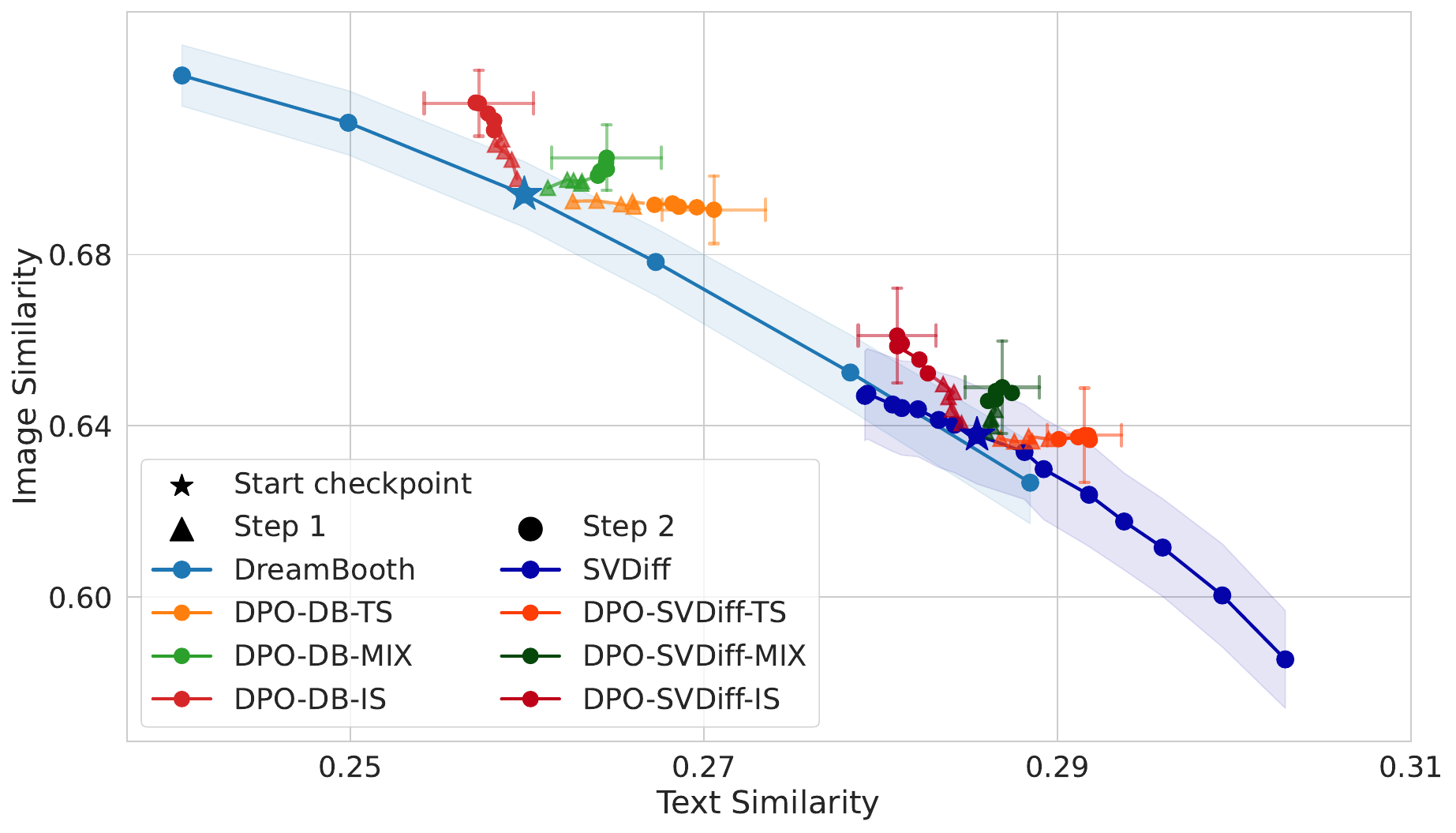}
      \captionof{figure}{SD2 results across DPO variants for long prompt setup. -TS setup improves prompt alignment, -IS enhances visual similarity, and -MIX balances both. All variants outperform baselines, enabling controllable trade-offs of TS and IS.}
      \label{fig:long:sd2}
  \end{minipage}
  \hfill
  \begin{minipage}{.40\textwidth}
      \centering
      \includegraphics[clip,width=\linewidth]{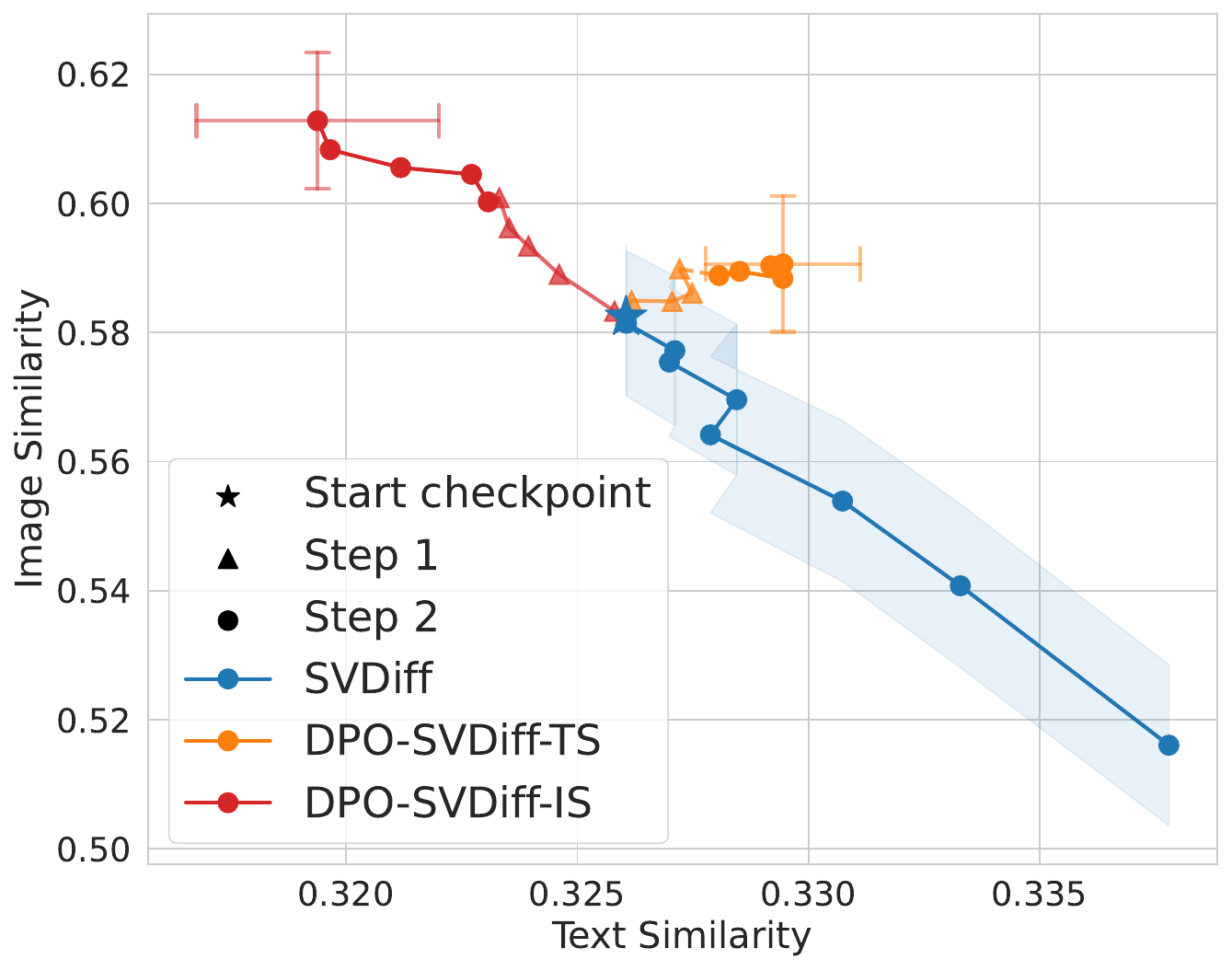}
      \captionof{figure}{Performance of DB-DPO on SDXL for long prompt setup. Improvements are consistent with SD2, demonstrating strong generalization to more capable base models.}
      \label{fig:long:sdxl}
  \end{minipage}
  \label{fig:long}
\end{figure*}

To evaluate the robustness of our method in more complex scenarios, we curated a set of $10$ long prompts for each of the \textit{live} and \textit{object} categories (see Table~\ref{tab:long-prompts}). These prompts are designed to be challenging, combining multiple scene modifications simultaneously. As shown in Figure~\ref{fig:long:sd2}, our method maintains strong performance, exhibiting behavior consistent with that observed on standard prompts. Furthermore, Figure~\ref{fig:long:sdxl} demonstrates that DB-DPO provides even better controllability when applied to more capable architectures like SDXL, enabling clear improvements along targeted directions.
Figure~\ref{fig:samples:long} provides qualitative comparisons, illustrating that the method consistently enhances prompt adherence and image fidelity relative to the baseline approach.

\begin{table}[b]
    \centering
    \begin{tabularx}{\linewidth}{p{0.45\linewidth}|p{0.45\linewidth}}
        \textbf{Live} & \textbf{Object} \\\hline
        [V*] in a chief outfit in a nostalgic kitchen filled with vintage furniture and scattered biscuit & a digital illustration of a [V*] on a windowsill in Tokyo at dusk, illuminated by neon city lights, using neon color palette \\\hline
        [V*] sitting on a windowsill in Tokyo at dusk, illuminated by neon city lights, using neon color palette & a sketch of a [V*] on a sofa in a cozy living room, rendered in warm tones \\\hline
        a vintage-style illustration of a [V*] sitting on a cobblestone street in Paris during a rainy evening, showcasing muted tones and soft grays & a watercolor painting of a [V*] on a wooden table in a sunny backyard, surrounded by flowers and butterflies \\\hline
        an anime drawing of a [V*] dressed in a superhero cape, soaring through the skies above a bustling city during a sunset & a [V*] floating in a bathtub filled with bubbles and illuminated by the warm glow of evening sunlight filtering through a nearby window \\\hline
        a cartoonish illustration of a [V*] dressed as a ballerina performing on a stage in the spotlight & a charcoal sketch of a giant [V*] surrounded by floating clouds during a starry night, where the moonlight creates an ethereal glow \\\hline
        oil painting of a [V*] in Seattle during a snowy full moon night & oil painting of a [V*] in Seattle during a snowy full moon night \\\hline
        a digital painting of a [V*] in a wizard's robe in a magical forest at midnight, accented with purples and sparkling silver tones & a drawing of a [V*] floating among stars in a cosmic landscape during a starry night with a spacecraft in the background \\\hline
        a drawing of a [V*] wearing a space helmet, floating among stars in a cosmic landscape during a starry night & a [V*] on a sandy beach next to the sand castle at the sunset with a floaing boat in the background \\\hline
        a [V*] in a detective outfit in a foggy London street during a rainy evening, using muted grays and blues & an anime drawing [V*] on top of a white rug in the forest with a small wooden house in the background \\\hline
        a [V*] wearing a pirate hat exploring a sandy beach at the sunset with a boat floating in the background & a vintage-style illustration of a [V*] on a cobblestone street in Paris during a rainy evening, showcasing muted tones and soft grays \\\hline
    \end{tabularx}
    \vspace{0.6em}
    \caption{Long prompts}
    \label{tab:long-prompts}
\end{table}

\FloatBarrier
\section{User Study details} \label{app:user}

\begin{figure}[h]
    \centering
    \includegraphics[width=\linewidth]{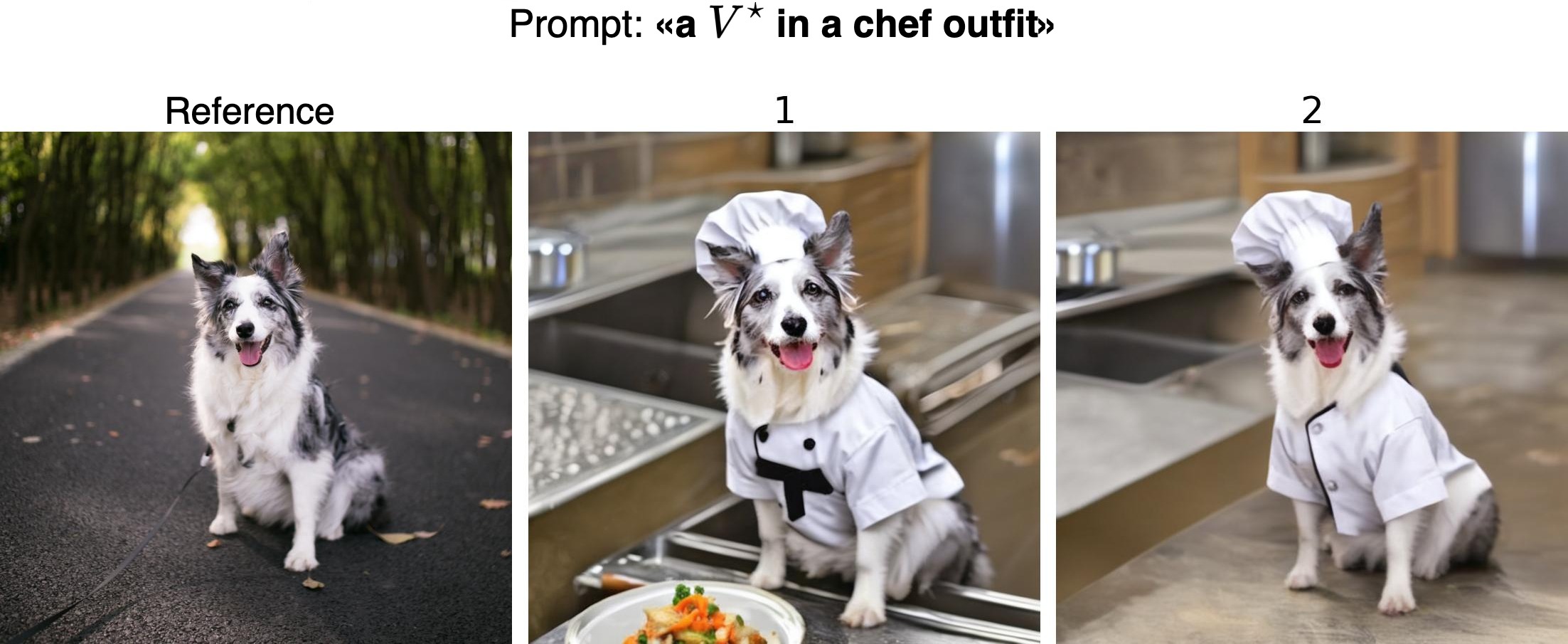}
    \caption{An example of a task in the user study}
    \label{fig:us:example}
\end{figure}

To validate our quantitative and qualitative observations, we conducted a User Study. We collected $6000$ responses from $10$ unique users for $2000$ unique pairs. An example task is shown in Figure~\ref{fig:us:example}. For each task, users were asked three questions to estimate the method in TS, IS, and All contexts: 1) \textit{"Which image is more consistent with the text prompt?"} 2) \textit{"Which image better represents the original image?"} 3) \textit{"Which image is generally better in terms of alignment
with the prompt and concept identity preservation?"} For each question, users selected one of three responses: \textit{"1"}, \text{"2"}, or
\textit{"Can't decide."}.

\FloatBarrier

\FloatBarrier
\section{Additional results} \label{app:more}

\begin{figure}[h]
    \centering
    \includegraphics[width=\linewidth]{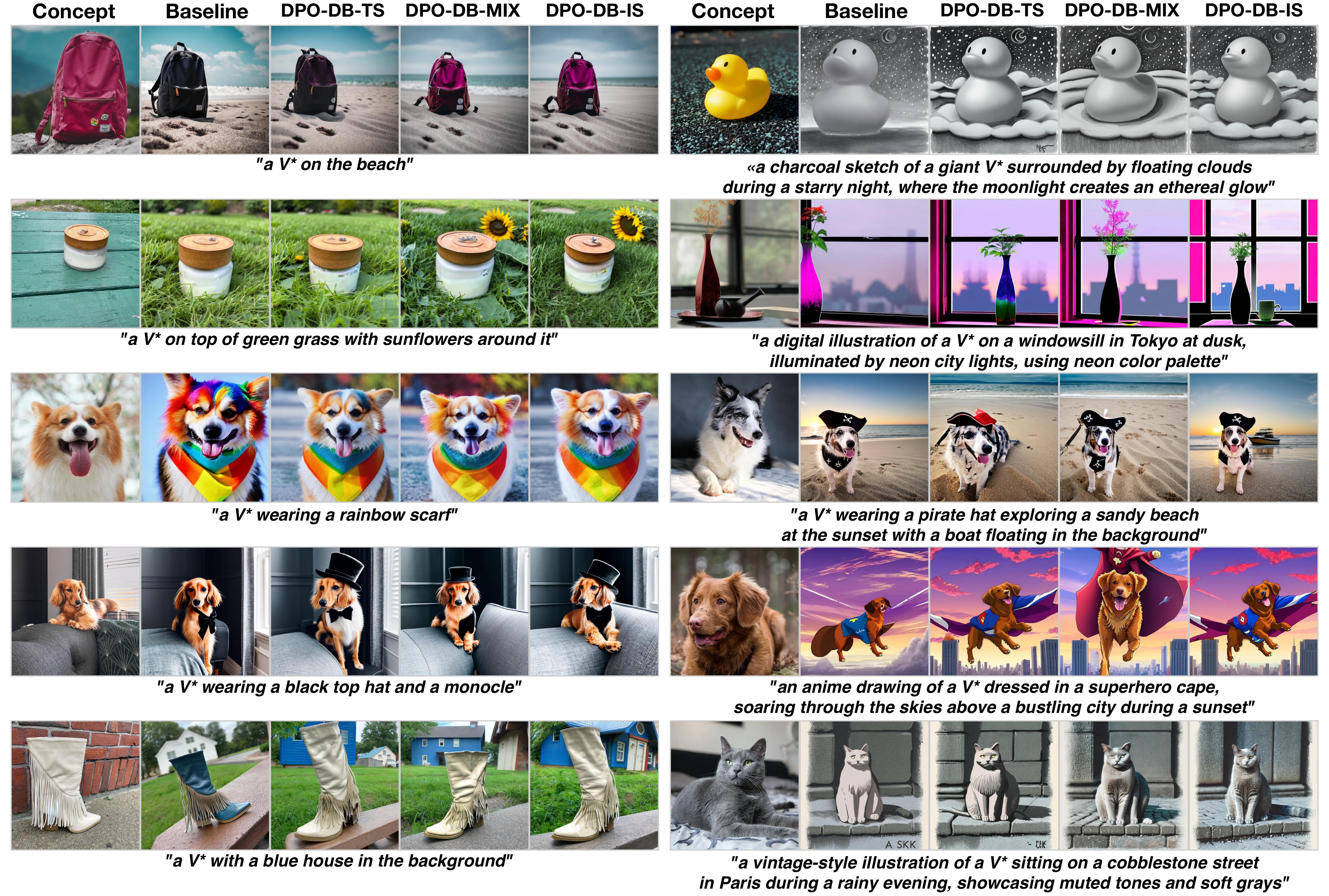}
    \caption{Additional qualitative examples for DPO-DB setup for standard and long prompts compared to the baseline.}
    \label{fig:extra:long:db}
\end{figure}

\begin{figure}[h]
    \centering
    \vspace{-0.8em}
    \includegraphics[width=\linewidth]{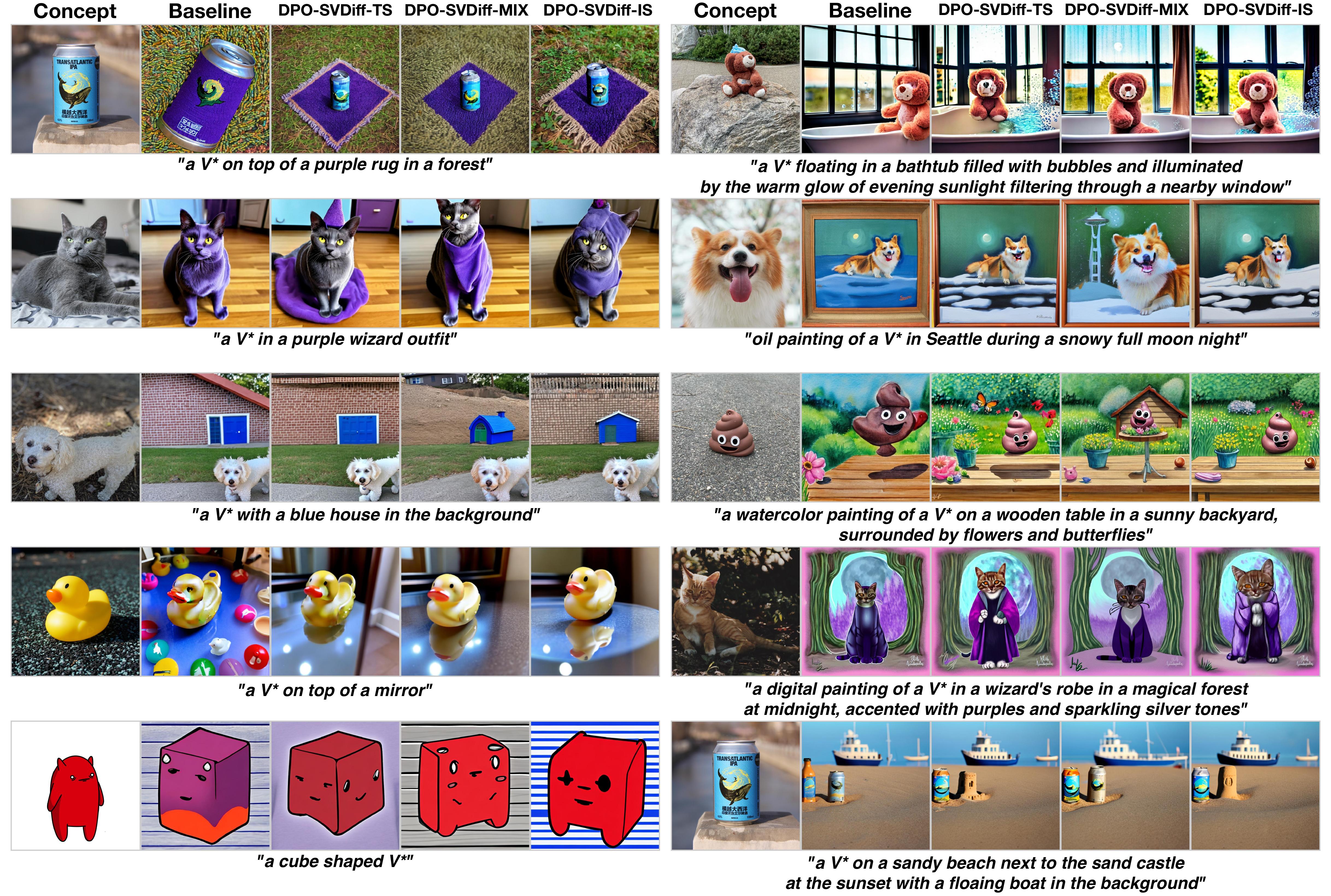}
    \caption{Additional qualitative examples for DPO-SVDiff setup for standard and long prompts compared to the baseline.}
    \label{fig:extra:long:svd}
\end{figure}

\begin{figure}[h]
    \centering
    \includegraphics[width=\linewidth]{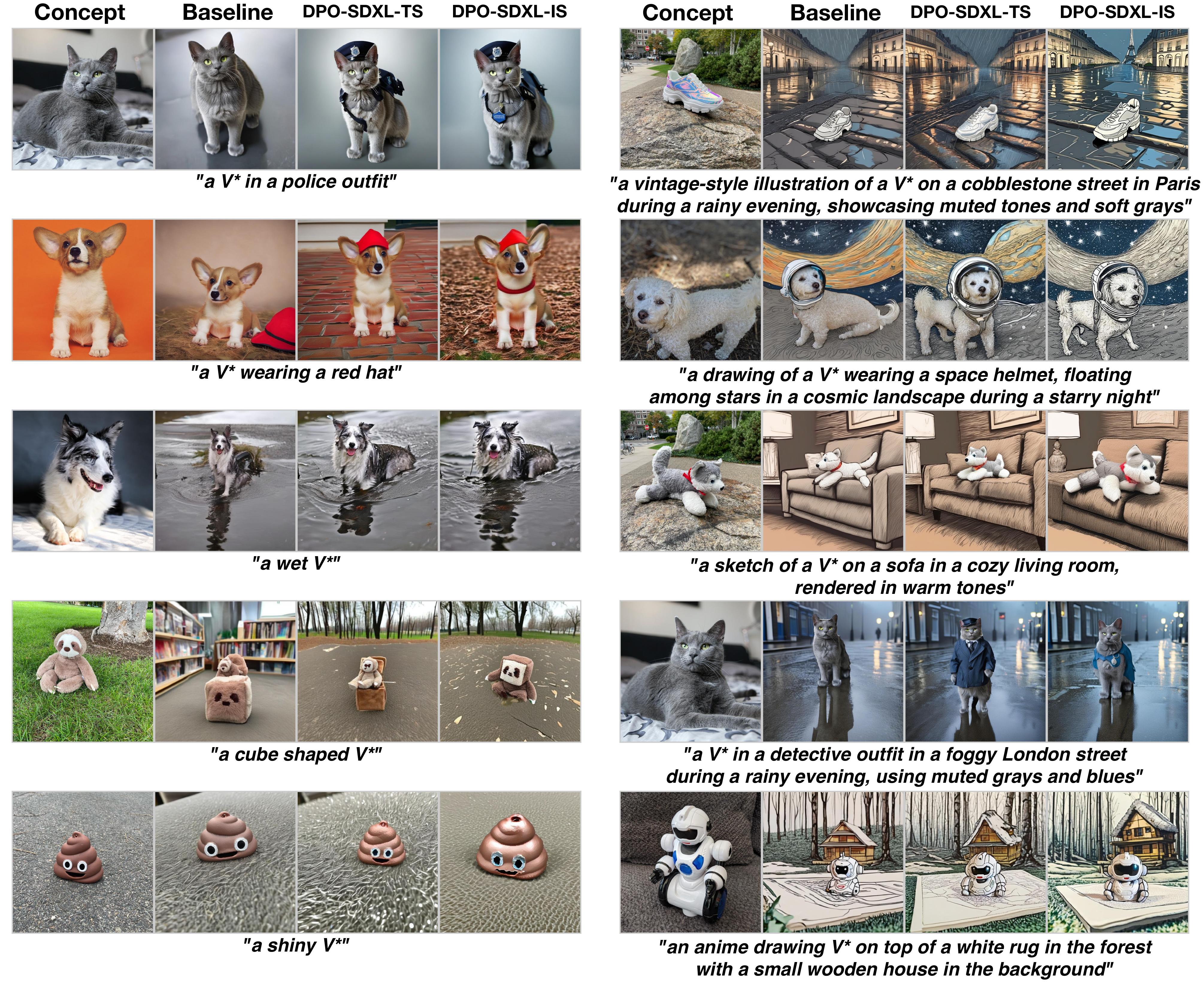}
    \caption{Additional qualitative examples for DPO-SDXL setup for standard and long prompts compared to the baseline.}
    \label{fig:extra:long:sdxl}
\end{figure}

Figures~\ref{fig:extra:long:db},~\ref{fig:extra:long:svd},~\ref{fig:extra:long:sdxl} provide additional qualitative comparisons. The key improvements achieved by our method include:
\begin{itemize}
    \item \textbf{Improved preservation of concept features} -- for instance, the \textit{color of a backpack} or the \textit{shape of a toy} is better retained, especially in the \textsc{-IS} setup.
    \item \textbf{Stronger alignment with textual prompts} -- examples include more accurate interpretations such as a \textit{"blue house in the background"} rendered as an actual house (not just a blue door), a \textit{"purple wizard outfit"} reflected in clothing (not merely color), or a \textit{"wet dog"} that visibly appears wet, rather than just surrounded by water. These improvements are particularly evident in the \textsc{-TS} and \textsc{-MIX} setups.
    \item \textbf{Addition of prompt-relevant details} -- our method generates elements often missed by the baseline, which is especially beneficial in long prompts that require complex scene construction. For example, the inclusion of the \textit{Seattle tower} in a prompt for \textit{dog}, or \textit{floating clouds} for \textit{duck\_toy}.
    \item \textbf{Fewer visual artifacts} -- outputs generally contain fewer distortions or inconsistencies. For example, the concept \textit{duck\_toy} appears correctly positioned \textit{"on top of a mirror"} with higher concept fidelity than in the baseline.
\end{itemize}

\FloatBarrier

\newpage

\end{document}